\documentclass{article}

    \PassOptionsToPackage{numbers}{natbib}
\usepackage[preprint]{neurips_2019}

\usepackage[utf8]{inputenc} % allow utf-8 input
\usepackage[T1]{fontenc}    % use 8-bit T1 fonts
\usepackage{hyperref}       % hyperlinks
\usepackage{url}            % simple URL typesetting
\usepackage{booktabs}       % professional-quality tables
\usepackage{amsfonts}       % blackboard math symbols
\usepackage{nicefrac}       % compact symbols for 1/2, etc.
\usepackage{microtype}      % microtypography
\usepackage{kotex}

\usepackage{graphicx}

\usepackage[english]{babel}

\usepackage{color}
\usepackage{xcolor}

\usepackage{algorithm}
\usepackage{algorithmicx}
\usepackage{algpseudocode}
\usepackage{subfigure}

\usepackage{amsmath,amssymb,amsfonts,amsthm,mathtools,bm}
\usepackage{enumitem}
\usepackage{multirow}
\usepackage{bm}
\usepackage{array}
\usepackage{color}
\usepackage{xcolor}
\usepackage{wrapfig}
\usepackage{multicol}
\usepackage{caption}
\usepackage{diagbox}
\usepackage{float}
\newcommand{\indep}{\rotatebox[origin=c]{90}{$\models$}}

\definecolor{customred}{RGB}{188, 50, 44}

\DeclareMathOperator*{\maximize}{maximize}

\definecolor{customred}{RGB}{192, 0, 0}

\title{Reliable Estimation of Individual Treatment Effect with Causal Information Bottleneck}

\author{%
Sungyub Kim$^{1}$, Yongsu Baek$^{1}$, Sung Ju Hwang$^{1,2}$, Eunho Yang$^{1,2}$\\
   KAIST$^{1}$, AItrics$^{2}$, South Korea\\
   \texttt{\{sungyub.kim, yongsubaek, sjhwang82, eunhoy\} @kaist.ac.kr} \\
}

\begin{document}

\maketitle

\begin{abstract}
  Estimating individual level treatment effects (ITE) from observational data is a challenging and important area in causal machine learning and is commonly considered in diverse mission-critical applications. In this paper, we propose an information theoretic approach in order to find more reliable representations for estimating ITE. We leverage the Information Bottleneck (IB) principle, which addresses the trade-off between conciseness and  predictive power of representation. With the introduction of an extended graphical model for \emph{causal} information bottleneck, we encourage the independence between the learned representation and the treatment type. 
  We also introduce an additional form of a regularizer from the perspective of understanding ITE in the semi-supervised learning framework to ensure more reliable representations. Experimental results show that our model achieves the state-of-the-art results and exhibits more reliable prediction performances with uncertainty information on real-world datasets.
  \end{abstract}

\section{Introduction}

% 1. (ITE 문제의 소개)\\
    Estimating individual-level treatment effect (ITE) , which infers causality between a treatment and an outcome from observational data, is one of the fundamental problems in machine learning and is essential in various applications such as healthcare~\cite{Shalit17,Glass13,Alaa17}, political science~\cite{Smith2005,LaLonde1986}, and online advertisement~\cite{Li16, Wang15} to name a few. In order to accurately deduce the causality between a treatment and an outcome, it is necessary to predict the outcome of each treatment correctly for each individual data point. However, in the usual (or almost all) observational studies, each data point is only allowed to receive one treatment out of several options, and the results for other unselected treatments do not even exist in the data. Hence, this ITE problem is connected to counterfactual questions \cite{johansson16} like ``If the patient received other prescriptions, would the patient's disease progress differently?''. A naive approach for this is to predict counterfactual outcomes that have not been observed, using the model trained only on observed factual data. One implicit assumption of such an inference is that the input distributions of the treatment and control groups (in case of two options) should be identical. It is known, however, that a bias may exist in the selection of treatments, unless the data are obtained in a rigorously randomized controlled trial, leading to a covariate shift in the ITE problem.
  
    To overcome this difficulty, previous works focus on a representation learning approach. A common central idea in this vein is to learn the feature extractor by which any discrepancy between the treatment and the control becomes smaller in the learned representation space than in the original covariate space. Almost all existing works based on this representation learning approach for ITE can be seen as a way of finding \emph{balanced and maximally expressive} representation that could be simultaneously used in the prediction of factual and counterfactual outcomes.

  %한다는 점에서 observational data의 selection bias에 대처하는 원칙을 가ㅣㅈ고 있다. 하지만 여기에는 representation이 observed covariate와 어떤 관계를 추구해야하는지 명확하지 않다는 점에서 missing point가 있다.
  % (예를 들어, observed covariate가 hidden confounding에는 없는 무의미한 noise를 포함하고 있을 경우 기존의 representation learning은 이를 적절하게 배제할 수 없다.) 
  
% 6.(missing point를 위한 Information Bottleneck approach)\\
  Motivated by the recent success of information bottleneck in several domains \cite{Alemi17, alemi2018, michael2018on}, in this paper we propose a \emph{causal information bottleneck (CIB)} that additionally pursue \emph{maximal compressiveness} between the representation for ITE prediction and the observed covariate, on top of previously considered \emph{balance and maximal expressiveness}.   
   % using an additional regularization, on top of previously considered \emph{balance and maximal expressiveness}, to pursue \emph{maximal compressiveness} between the representation for ITE prediction and the observed covariate, via the information bottleneck principle \cite{Tishby99}. %{\color{red} revise: Consider the situation where there is a hidden confounder affecting both the covariate and outcome as in most problems. In this case, directly modeling  .
  %
  %, and the covariate $X$ is just a noisy proxy of $Z$
  %the information bottleneck principle is important in the context of causal inference, as observed covariates are not hidden confounders but are instead 'proxies' of it.~\cite{Louizos17}. It is a well-known problem that using a covariate as a hidden confounder can cause bias\cite{pearl2012, kuroki2014}}.
%   {\color{red} {\bf related work으로 이동}: ~\cite{Louizos17}에서는 이를 극복하기 위해 latent variable를 variational inference를 통해 학습시킬 것을 제안했으나, auto-encoding 구조에서 사용되는 reconstruction loss 때문에 X와 Z 사이의 compression에는 한계가 있다.} 
%    
    By learning with information bottleneck principles \cite{Tishby99}, not only can one learn an ideal representation that can better predict the counterfactual outcomes, but one can also gain additional (but more important in some sense) effects. Specifically, earlier work \cite{Alemi17} showed that models learned with information bottleneck objective are more robust to overfitting and adversarial attacks than models learned through other forms of objectives or regularizations. This occurs because the representation found through information bottleneck tries to ignore as much information as possible from the covariate and hence idiosyncratic perturbations cannot easily pass through an information bottleneck. In addition to improving generalization, models learned by IB principals are empirically shown to be well calibrated providing meaningful uncertainty about their predictions \cite{alemi2018}. In the same context, one can expect that the proposed CIB naturally inherits such benefits of IB. Toward this, we experimentally confirm that CIB yields significantly better results against baselines on the scenario where the models are allowed to say ``I don't know'' on instances that are different from the training data and hence they are not sure. It is one of the most important desiderata to provide accurate uncertainty on their predictions since most tasks that consider the counterfactual inference are irreversible and mission-critical problems with serious consequences.

\paragraph{Main contributions.} The main contributions of this paper are as follows:
\begin{itemize}
    \item We introduce a graphical model for causal inference on which we can apply the information bottleneck (IB) principle for reliable individual treatment effect (ITE) estimations. The proposed principle is formulated as a simple optimization problem that can be efficiently solved with stochastic gradient methods \cite{Kingma2015} and a reparameterization trick \cite{Kingma2014}.
    \item Along with the standard information bottleneck, we propose to use two additional regularizations, one that emerges naturally by the graph structure and another inspired by semi-supervised learning literature, to help learning representations for ITE.
    \item We validate our model on three standard benchmark datasets and show that our model significantly outperforms with well-calibrated uncertainty for the scenario where the models are allowed to say ``I don't know''.
\end{itemize}
    
% 9. (본 논문의 contribution을 추가적으로 쓸 수 있으면 좋을 것 같음.)

\section{Background and Setup}

\paragraph{Individual Treatment Estimation (ITE)}
% 1. (ITE의 fomal한 정의)\\
    Here, we denote the set of possible covariates by $\mathcal{X}$ and the set of possible outcomes by $\mathcal{Y}$. We focus on a binary treatment scenario that selects one among the treat/control options, just for simplicity.
    % {\color{red} 일반적인 케이스로 익스텐션 가능? 가능하다면 simplicity를 위해서 binary case에 집중한다고 언급할 것}. 
    For convenience of the notation, we use $0$ to denote the control and $1$ to denote the treat, that is, the space of treatment is $\mathcal{T} = \{0,1\}$. For example, let $x_i \in \mathcal{X}$ be a covariate describing the features of patient $i$. If the patient chooses to receive some treatment, $t_i = 1$, otherwise $t_i = 0$. The actual patient progression variable (e.g. blood pressure, blood sugar) after selecting the medical care becomes $y_i^{F} \in \mathcal{Y}$. Such a set of factual data can be expressed as $\{(x_i, t_i, y_i ^ F)\}_{i = 1}^{n}$ for $n$ patients. Similarly, the counterfactual data can be expressed as $\{(x_i, 1- t_i, y_i^{CF})\}_ {i = 1}^{n}$, however there is no $ y_i^{CF}$ value available in the observational data. The individual treatment effect (ITE) that we want to estimate is a conditional expected difference between the treatment outcome $ y_i^{(1)} $ and the control outcome $ y_i^{(0)} $ given covariate $ x_i $,
    %\begin{align*}
        $\tau(x_i) = \mathbb{E}\big[ y_i^{(1)} - y_i^{(0)} | x_i \big]$.
    %\end{align*}
    A commonly used metric to evaluate the performance of ITE estimators is the \textit{expected Precision in Estimation of Heterogeneous Effect}~(PEHE)~\cite{hill2011bayesian}, which is defined as 
    %\begin{equation}
        $\epsilon_{PEHE} = \mathbb{E}_{x}\big[\big(\tau(x_i) - \hat{\tau}(x_i)\big)^2\big]$. Here,  $\hat{\tau}(x_i)$ is the predicted difference by the model. 
    %\end{equation}

% 2. (Potential Outcome model의 가정들, overlap, unconfoundedness)\\
    As several works on causal inference including \cite{johansson16, Shalit17, yao2018} did for justifiability of counterfactual prediction, we also assume throughout the paper the following two conditions of the Rubin-Neyman causal model \cite{Rubin1974, Shalit17}:
    \begin{itemize}
        \item (Overlap) $\mathbb{P}[t|x] \in (0,1) ,\forall x \in \mathcal{X}, \forall t \in \mathcal{T}$,
        \item (Conditional ignorability) $Y \indep T | X$.
    \end{itemize}
    The overlap assumption here requires that all patients have a strictly greater than $0$ probability of receiving all treatments. This assumption is necessary to ensure that each patient has both possibilities of being treated or controlled so that so that ITE can be meaningfully estimated. The assumption of conditional ignorability implies that the covariate has sufficient information about the common effects of treatment decisions and outcomes. Therefore, it is also referred to as an unconfoundedness assumption \cite{Rubin2005}. This unconfoundedness assumption is necessary to make the estimation of ITE identifiable~\cite{Rubin15}.

\begin{figure}[t]
	\centering
    \subfigure[]{\includegraphics[width=0.25\linewidth,trim={0 1.2cm 0 0 },clip]{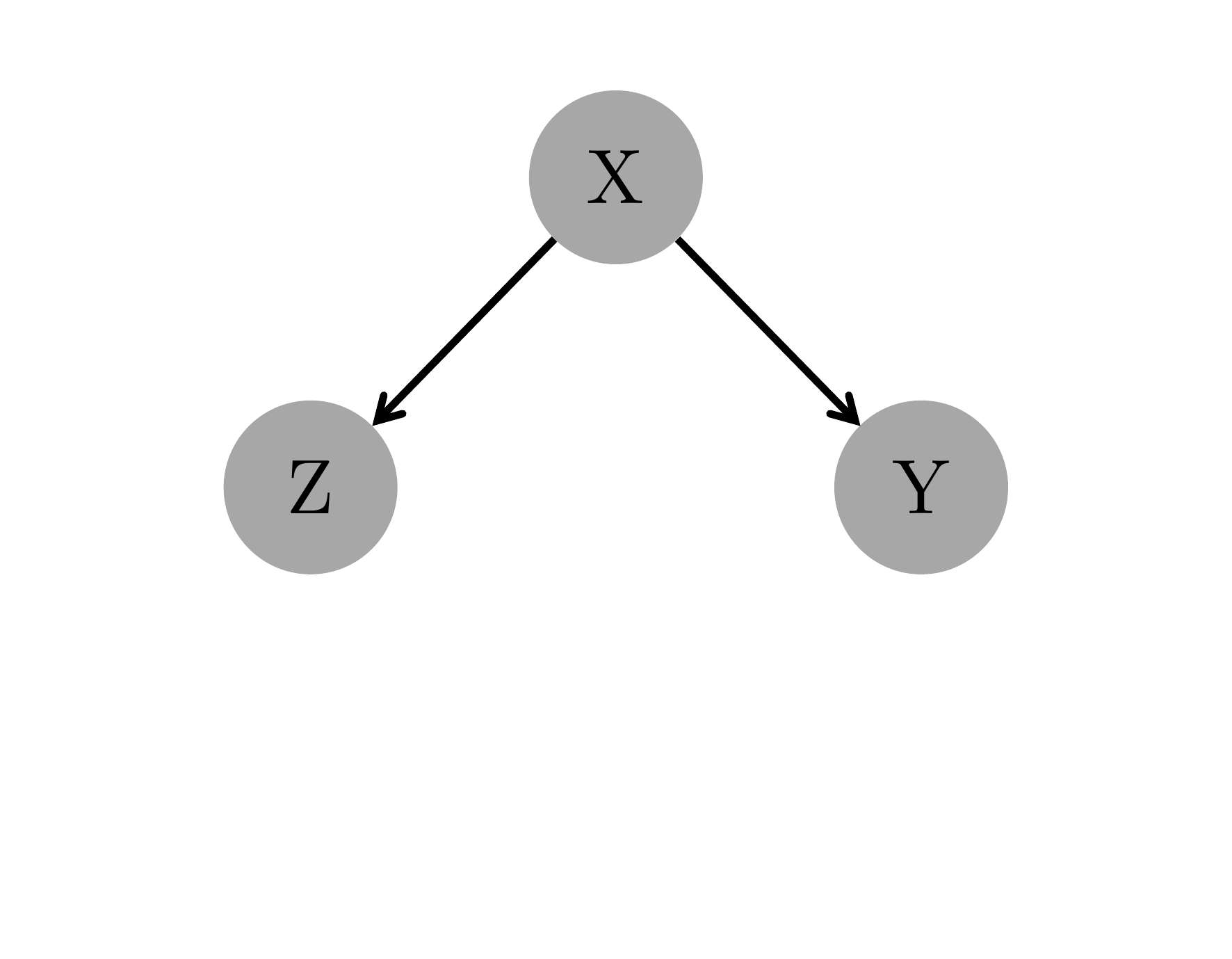}}
 	\subfigure[]{\includegraphics[width=0.25\linewidth,trim={0 1.2cm 0 0 },clip]{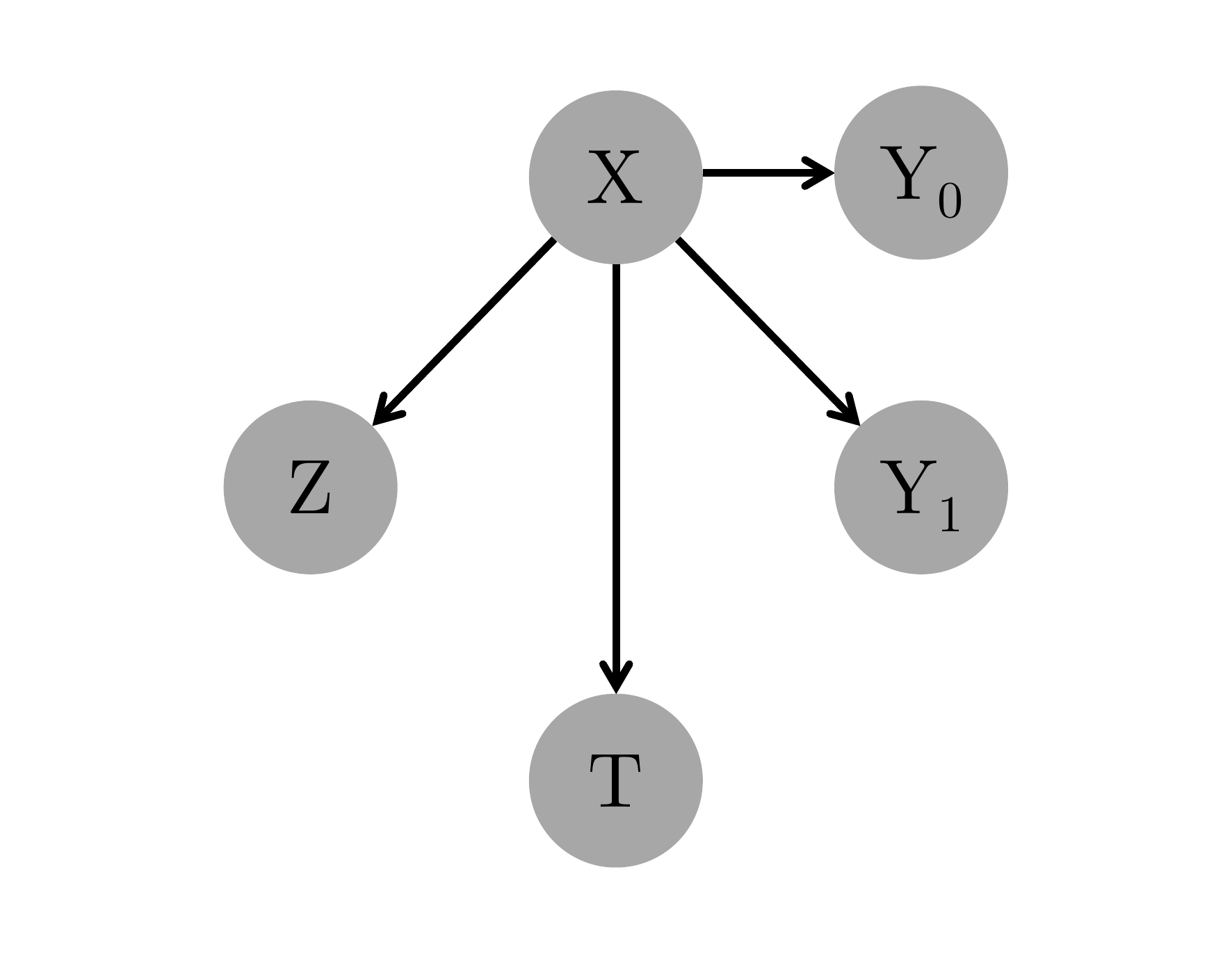}}
 	\subfigure[]{\includegraphics[width=0.25\linewidth,trim={0 1.2cm 0 0 },clip]{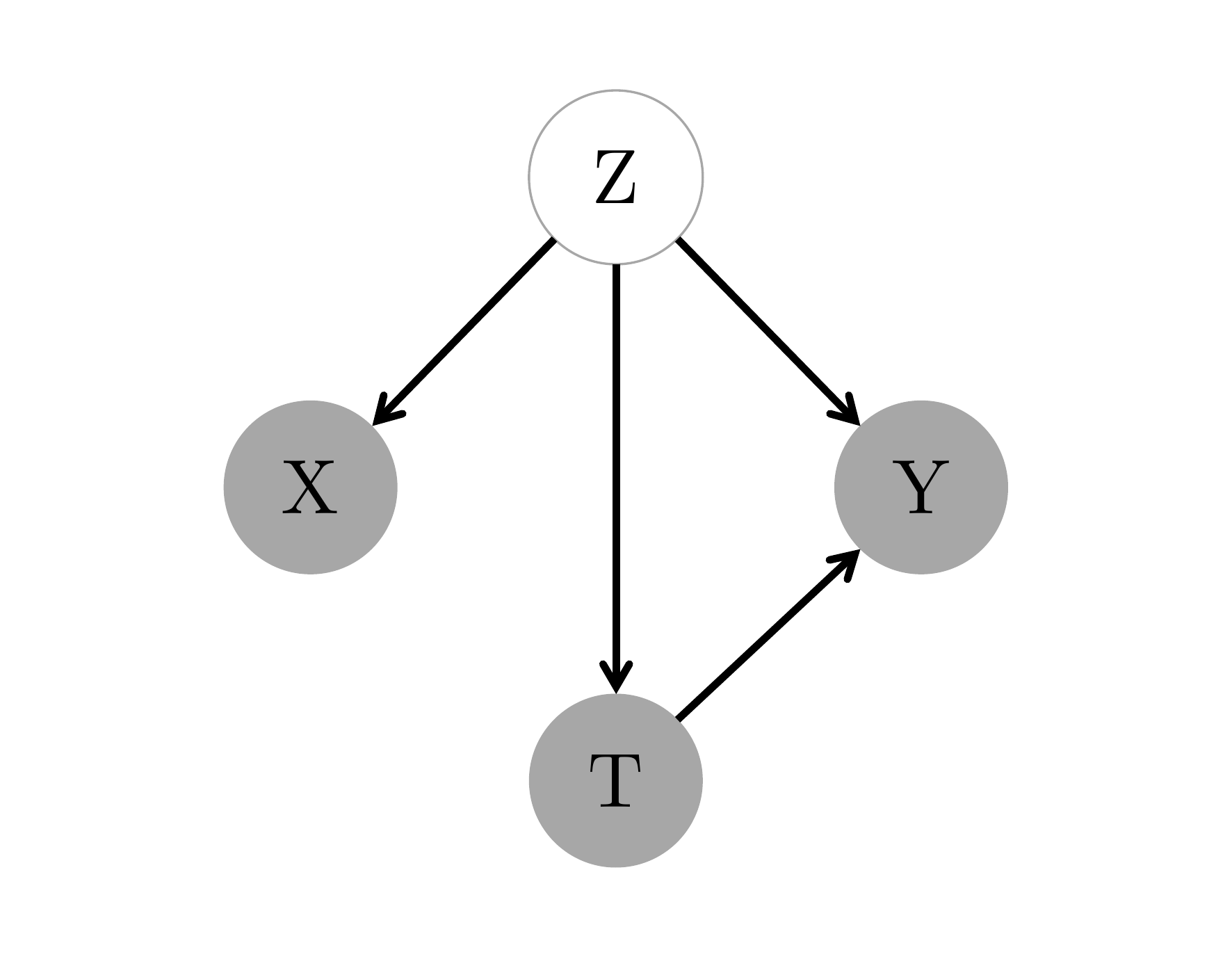}} % x와 z 위치 change?
     \vspace{-.2cm}\caption{(a) Graphical model of classical information bottleneck (b) Graphical model of causal information bottleneck (c) Graphical model of CEVAE~\cite{Louizos17} } 
 	\label{fig:graphical_model}
\end{figure}

\paragraph{Information Bottleneck (IB) principle}    
    The \textit{information bottleneck principle} was first proposed in \cite{Tishby99}. The trade-off between \textit{maximal expressiveness} and \textit{maximal compressiveness} in the information bottleneck theory is expressed by the following expression, which contains two mutual information terms $\mathcal{I}(\cdot, \cdot)$: one between input $X$ and representation $Z$ and the other between representation $Z$ and output $Y$:
  \begin{align*}
      \maximize \mathcal{I}(Y;Z) - \beta \, \mathcal{I}(X;Z).
  \end{align*}
  The classical information bottleneck assumes the graphical model shown in Figure \ref{fig:graphical_model}(a) and computes $p_{\phi}(z|x)$ by sampling from stochastic encoder given input. In general, it is known that the information bottleneck principle is difficult to pursue since a precise estimation of the mutual information is intractable. However, recent works such as those focusing on variational approximation \cite{Alemi17} and adversarial learning \cite{Belghazi18} proposed efficient techniques to approximate mutual information and confirmed that information bottleneck is an effective regularization methodology robust to observation noise, hence providing better generalization performance and resilience against adversarial attacks and so on. 
  
\section{Causal information bottleneck (CIB)}
 It is important to note that even under the conditional ignorability above, the covariate $X$ might have information irrelevant to both $Y$ and $T$ in many real problems. Such \emph{noise} in covariates cause problems such as overfitting. %when estimating an individual treatment based on factual data. The overfitting problem is important in relation to causal inference for two reasons. 
    Since in the main application fields of causal inference such as health care and political science, it is hard to collect huge number of training data, and moreover they are usually irreversible mission-critical problems, we need a systematic way of removing such noise present in the covariate and having a reliable representation for predictions.
    
    %cannot easily obtain data compared to the fields of computer vision and natural language processing, and the possibility of overfitting during learning process is high. Second, misprediction in these applications is mission-critical. Owing to the opportunity cost and the irreversible nature of the prescription and decision-making using causal inference, incorrect estimations can cause a large losses.

%\subsection{Causal Information Bottleneck}

% 1. (Causal IB의 각 gradphical model의 component들이 필요한 이유 설명. covariate, representation, treatment별 outcome node들 도입. 이들이 만족해야 하는 조건들 개조식으로 언급.\\
% 1. representation과 outcome의 MI maximize\\
% 2. representation과 covariate의 MI minimize\\
% 3. representation과 outcome들의 conditional independence)\\
    Toward this, in this section we propose an information bottleneck framework for causal inference. We extend the graphical model in Figure \ref{fig:graphical_model}(a) into the form of Figure \ref{fig:graphical_model}(b) to incorporate variables on treatment and factual/counterfactual outcomes. %In this graph structure, representation $Z$ is conditionally independent with outcomes  $Y_0$ and $Y_1$ when the covariate $X$ is given. 
    Unlike the graphical models to capture data generation process with hidden \emph{confounder} $Z$ (for example, CEVAE \cite{Louizos17} in Figure \ref{fig:graphical_model}(c)), the graphical model of CIB directly extends the information bottleneck to causal inference task, to find some good \emph{representation} in the discriminative setting.
    %, explicitly modeling the outcome $ Y_0, Y_1 $ rather than modeling outcome $ Y $, with only one observed by treatment $T$ .  
    Note that treatment $ T $ and outcome $ Y $ are conditionally independent given covariate $ X $, which is a sufficient condition for the conditional ignorability of the Rubin-Neyman causal model. Under the graphical model shown in Figure\ref{fig:graphical_model}(b), we propose the information bottleneck principle for the causal model:
    \begin{itemize}
        \item \textbf{(Maximal Expressiveness)} representation $Z$ should have high mutual information with treatment outcome $ Y_0, Y_1 $.
        \item \textbf{(Maximal Compressiveness)} representation $Z$ should have low mutual information with covariate $ X $.
    \end{itemize}
    These principles can be expressed as the following optimization problem, with a tunable Lagrange multiplier $\beta$:
    \begin{align}\label{EqnCIB}
      \maximize \mathcal{I}(Y_0;Z) + \mathcal{I}(Y_1;Z) - \beta \mathcal{I}(X;Z).
    \end{align}
    
% 2. (위에서 언급한 첫번째 수식의 variational inference 수식 유도. 여기서는 기존의 IB와 다르게 각각의 outcome에 대해 variational inference가 유도됨을 언급. 또한, graphical model의 가정을 사용하는 부분 강조)\\

    The first term in the objective \eqref{EqnCIB}, \textit{maximal expressiveness} of $ Z $ for control outcome $ Y_0 $, can be calculated by the following variational approximation \cite{Alemi17}:
    \begin{align}
        \mathcal{I}(Y_0; Z)&= \displaystyle\int\int p(y_0,z) \log \frac{p(y_0,z)}{p(y_0)p(z)} dy_0dz = \displaystyle\int\int p(y_0,z) \log \frac{p(y_0|z)}{p(y_0)} dy_0dz \nonumber\\
        &\ge \displaystyle\int\int p(y_0,z) \log \frac{q_{\theta}(y_0|z)}{p(y_0)} dy_0dz\label{max-exp-va}\\
        &= \displaystyle\int\int p(y_0,z) \log q_{\theta}(y_0|z) dy_0dz + \mathcal{H}(y_0) \label{max-exp-va-calc}
    \end{align}
    where $q_{\theta}(y|z)$ is a variational approximation of the conditional distribution $p(y|z)$ and $\mathcal{H}(y_0)$ is the entropy of the random variable $y_0$. The inequality (\ref{max-exp-va}) follows from the non-negativity property of the Kullback Leibler (KL) divergence:
    \begin{align*}%\label{non-neg-kl}
        D_{KL}(p(y_0|z)\|q_{\theta}(y_0|z)) \ge 0 \Longrightarrow \displaystyle\int p(y_0|z) \log p(y_0|z)  dy_0 \ge \displaystyle\int  p(y_0|z) \log q_{\theta}(y_0|z) dy_0, \forall z.
    \end{align*}
    Now (\ref{max-exp-va-calc}) can be calculated as follows with a stochastic encoder $p_{\phi}(z|x)$:
    \begin{equation}\label{variational-information}
        \displaystyle\int\int p(y_0,z) \log q_{\theta}(y_0|z) dy_0dz = \displaystyle\int\int  p(y_0|x) p_{\phi}(z|x) p(x) \log q_{\theta}(y_0|z) dy_0dzdx.
    \end{equation}
    This follows from the fact that $p(y_0,z) = \int p(y_0, z, x) dx = \int p(y_0, z|x) p(x)dx = \int p(y_0|x)p_{\phi}(z|x) p(x) dx$ under the graph structure Figure \ref{fig:graphical_model}(b).  
    
    Here, additional difficulty arises in computing \eqref{variational-information} due to the nature of observational data; we must marginalize $p(y_0 | x)$ over the entire population $p(x)$, but we know $y_0$ only for the conditional population $ p (x| t = 0) $.
    To overcome this issue, we can further derive \eqref{variational-information} under the overlap assumption $p(t = 0 | x) \in (0,1)$ above:
    \begin{align*}
        &\displaystyle\int\int p(y_0|x) p_{\phi}(z|x) p(x) \log q_{\theta}(y_0|z)  dy_0dzdx\\
        =& \displaystyle\int\int p(y_0|x) p_{\phi}(z|x) \frac{p(x)}{p(x|t=0)}p(x|t=0)\log q_{\theta}(y_0|z)  dy_0dzdx\\
        =& \displaystyle\int\int p(y_0|x) p_{\phi}(z|x) \frac{p(t=0)}{p(t=0|x)}p(x|t=0)  \log q_{\theta}(y_0|z)  dy_0dzdx.
    \end{align*}
    %Note that $ p (t = 0 | x) \in (0,1) $ as assumed by the unconfoundedness assumption, and it is difficult to determine the distribution precisely. Therefore, 
    At this point we introduce the score classifier $ s_{\nu} $ to estimate the probability of treatment from data and define the loss:
    \begin{align*}
         \mathcal{L}_{0}(\phi, \theta) {:=} \displaystyle\int\int p(y_0|x) p_{\phi}(z|x) \frac{p(t=0)}{s_{\nu}(t=0|x)}p(x|t=0) \log q_{\theta}(y_0|z) dy_0dzdx.
    \end{align*}
    We can derive the variational bound for the mutual information between $Z$ and $Y_1$ in a similar way.

% 3. (위에서 언급한 두번째 수식은 기존의 derive와 동일함을 언급. 결과적으로 variational marginal을 사용하는데 우리는 이 marginal을 학습시키며 이것이 regularization은 flexible하게 해줌을 언급)

    On the other hand, the last term in \eqref{EqnCIB}, the \emph{maximal compressiveness}, can be computed as a seamless extension of variation approximation for standard IB \cite{Alemi17}: $\mathcal{I}(X;Z) \le \displaystyle\int\int p(x,z) \log p_{\phi}(z|x) dxdz - \int p(z) \log r_{\psi}(z) dz {=:} \mathcal{L}_{C}(\phi, \psi)$
    %\begin{align}
      %\mathcal{I}(X;Z) &= \displaystyle\int\int p(x,z) \log \frac{p(x,z)}{p(x)p(z)} dxdz \\
      %&= \displaystyle\int\int p(x,z) \log p_{\phi}(z|x) dxdz - \int p(z) \log p(z) dz\\
      %&\le \displaystyle\int\int p(x,z) \log p_{\phi}(z|x) dxdz - \int p(z) \log r_{\psi}(z) dz.
    %\end{align}
    where $r_{\psi}(z)$ is a variational approximation of $p(z)$. %As in equation (\ref{non-neg-kl}), the following can be derived from the property of KL divergence\cite{Alemi17}.
    %\begin{align*}
    %    D_{KL}(p(z)\| r_{\psi}(z)) \ge 0 \Longrightarrow \displaystyle\int p(z) \log p(z)  dz \ge \displaystyle\int  p(z) \log r_{\psi}(z) dz .
    %\end{align*}
    %Note that the variational approximation $ r_{\psi}(z)$ can be any distribution, that the representation marginal $p(z)$ is absolutely continuous distribution with respect to. For that degree of freedom of $ r_{\psi}(z) $, \cite{Alemi17} set $r_{\psi} (z) $ as the standard normal distribution, while in \cite{alemi2018} this is set as a learnable mixture of Gaussian. In this paper, we choose diagonal Gaussian for learning.\\
    
% 4. (위에서 언급한 세 번째 수식은 graphical model 상에서 성립하는 가정이고 실제 학습에서는 만족할 이유가 없음을 언급. 따라서 우리는 이를 만족시키기 위해 추가적인 정규화가 필요함을 언급.)

    \begin{figure}
        \centering
        \includegraphics[width=1.0\linewidth, clip]{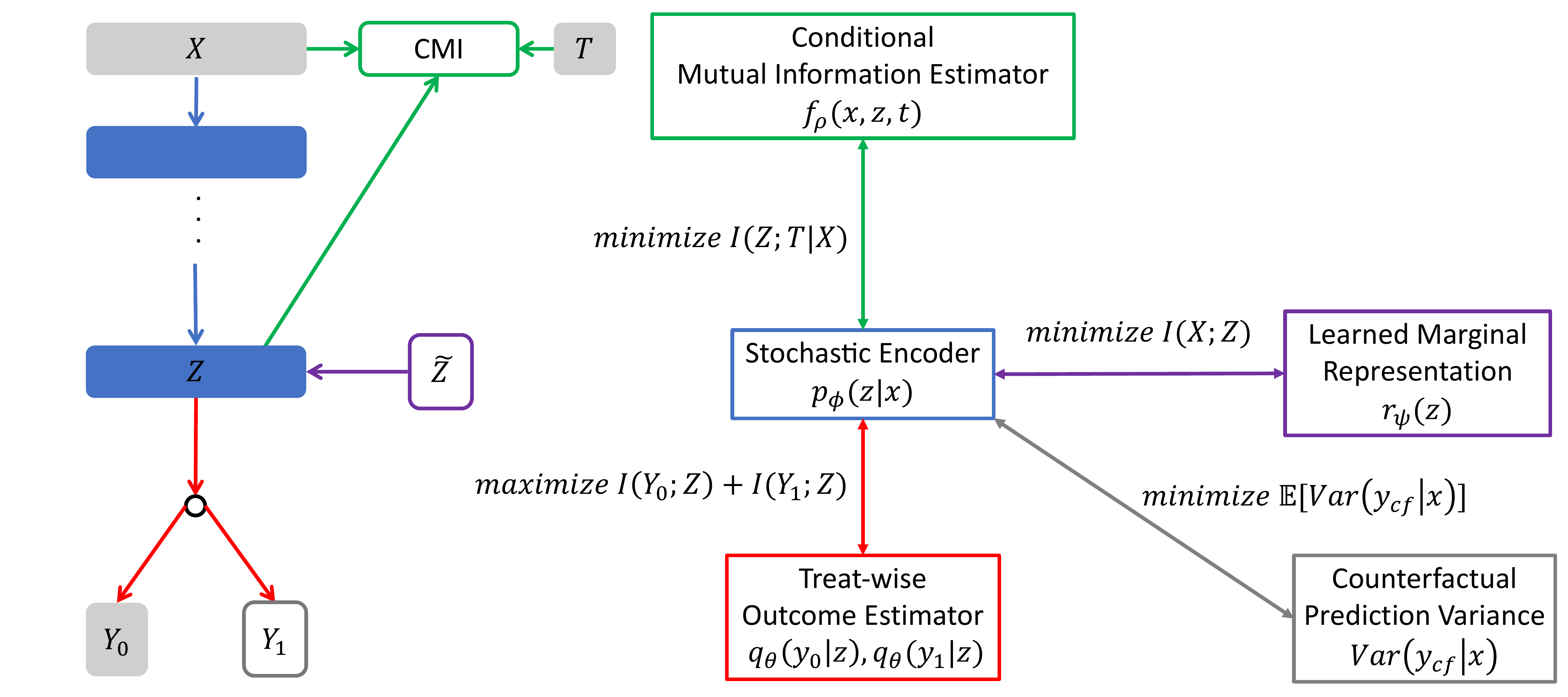}
        \caption{Overall architecture of Causal Information Bottleneck~(CIB). }%We skipped score classifier $s_{\nu}$ for simplicity.}
	    \label{fig:network_architecure}
    \end{figure}
    
    Meanwhile, the graphical model of Figure \ref{fig:graphical_model}(b) assumes that representation $Z$ is conditionally independent of the treatment $T$ given covariate $X$. Without additional constraint, this assumption can not be satisfied with a stochastic encoder $p_{\phi}(z|x)$. Hence, we introduce a novel technique referred to as mutual information guided disentangling that will be explained in the next subsection.  Furthermore, the ITE estimation where we are asked to predict unlabeled counterfactual outcomes, may be understood under the  semi-supervised learning framework. %우리의 causal information bottleneck은 stochastic encoder를 사용하여 predictive distribution을 표현할 수 있기 때문에, 
    Based on this interpretation, we apply predictive variance regularization, a popular method in semi-supervised learning \cite{jean2018semi}, for counterfactual prediction. 

\paragraph{Mutual Information Guided Disentangled Representation (MIGDR)}

% 1. (Mutual information의 정의와 이를 추정하는 것의 어려움에 대한 언급)\\
    %As mentioned above, since accurately estimating  mutual information is intractable, \cite{chen2016infogan, Alemi17, Belghazi18} proposed the approximate estimation of mutual information. In the present paper, we will apply regularization using the Donsker-Varadhan representation-based methodology\cite{MD1983} 
    In order to encourage $Z \perp\mkern-10mu\perp T |X$ according to our graphical structure, we minimize the conditional mutual information between them,
    \begin{align*}
        \mathcal{I}(Z;T|X) = \mathbb{E}_{p(x)}\big[D_{KL}(p(z,t|x)\|p_{\phi}(z|x)p(t|x))\big].
    \end{align*}
    Since applying the variational approximation technique as above is not trivial in this case due to the conditional expectation, we resort to the Donsker-Varadhan representation-based methodology \cite{MD1983}. According to the Donsker-Varadhan representation, the above KL divergence can be expressed as follows \cite{Belghazi18},
    \begin{align*}
    % \centering
        \mathbb{E}_{p(x)}\big[D_{KL}(p(z,t|x)\|p_{\phi}(z|x)p(t|x))\big]
        =\mathbb{E}_{p(x)}\Big[\sup_{f} \mathbb{E}_{p(z,t|x)}[f] - \log \mathbb{E}_{p_{\phi}(z|x)p(t|x)}[e^{f}]\Big]
    \end{align*}
    where $f : \mathcal{X} \times \mathcal{Z} \times \mathcal{T} \rightarrow \mathbb{R}$ is a scalar-valued function whose two expectations are finite. As proposed in \cite{Belghazi18}, this can be approximated efficiently by estimating auxiliary neural network $f_\rho$ with parameter $\rho$. This additional network discriminate between samples from joint distribution $p(z,t|x)$ and samples from product distribution $p(z|x)p(t|x)$. From $f_\rho$, one can efficiently minimize conditional mutual information between representation $Z$ and target label $T$ by maximizing
    \begin{equation}
        \mathcal{L}_{M}(\phi, \rho) {:=}  -\mathbb{E}_{p(x)}\big[ \mathbb{E}_{p(z,t|x)}[f_{\rho}(x,z,t) ] \big] .
    \end{equation}
    More details are provided in the appendix.

\paragraph{Counterfactual Predictive Variance Regularization (CPVR)}

% 1. (CMGP, SSDKL의 연구결과를 소개. 여기에서 일관적으로 언급하는 unlabelled data에 대한 variance regularization을 언급)\\
    Predictive variance regularization, a method commonly used for semi-supervised learning, adds variance minimization to the predictive distribution of unlabeled data as a regularization term \cite{jean2018semi}. In the regularized Bayesian framework, this method includes the intuition that an unlabeled sample is close to the labeled sample in the representation space \cite{Alaa17, jean2018semi}. In this paper, we provide an inductive bias so that counterfactual predictions generated by stochastic encoder are consistent to each other.
    \begin{align*}
        \mathcal{L}_{V}(\phi, \theta) {:=}  -\mathbb{E}_{p(x)}\big[ \text{Var}_{q_{\theta}(y|x)}[y^{CF}|x] \big].
    \end{align*}
    The predictive variance regularization is similar to the nearest neighbor methodology frequently used in counterfactual predictions \cite{johansson16, chang2017informative, li2017matching} in that both  provide an approximate target for counterfactual data. However, predictive variance regularization uses the predictions of learned networks, while nearest based methods propagate factual data to unlabelled counterfactual outcomes. The nearest neighbor method can be a good approximation when the reference data is sufficient, but the predictive variance regularization can provide better inductive bias if there is little data to calculate nearest neighbors. We use the gradient information of CPVR only to update the parameters of stochastic encoder as in \cite{Alaa17, jean2018semi}.
    
    Overall, the final objective of CIB is described as (with graphical description in Figure \ref{fig:network_architecure})
    \begin{align*}
        \maximize_{\theta, \phi, \psi} & \ \mathcal{L}(\theta, \phi, \psi) {:=}  \mathcal{L}_{0}(\phi, \theta) + \mathcal{L}_{1}(\phi, \theta) - \beta \cdot \mathcal{L}_{C}(\phi, \psi) + \lambda_{M}\cdot \mathcal{L}_{M}(\phi, \rho) + \lambda_{V}\cdot\mathcal{L}_{V}(\phi)
    \end{align*}
    where $\mathcal{L}_{M}(\phi, \rho)$ involves the following optimization problem
    \begin{align*}
        \maximize_{\rho}  \ \mathbb{E}_{p(x)}\Big[\sup_{f} \mathbb{E}_{p(z,t|x)}[f] - \log \mathbb{E}_{p_{\phi}(z|x)p(t|x)}[e^{f}]\Big].
    \end{align*}

\section{Related Works}

\paragraph{Balanced representation for causal inference}

% 1. Global Representation을 Balancing하는 논문들. BNN, CFR\\
    The idea of balancing the representation between heterogeneous samples is a commonly used idea in the field of unsupervised domain adaptation \cite{UDA15, tzeng2017adversarial, hoffman18}. This idea comes from the concept of $ \mathcal{H}$-divergence in \cite{ben2010theory}. It is proposed in \cite{johansson16, Shalit17} that this idea can be applied to counterfactual inference. %\cite{johansson16} showed that representation balancing between factual and counterfactual data can minimize the risk of counterfactaul error, and \cite{Shalit17} showed that representation balancing between control and treat data is also related to generalization error of counterfactual error. They showed that representation balancing idea is not only important theoretically but also critical to empirical performance. Representation balancing can be understood as regularizing $T$ and $Z$ marginally independent in that the distribution of representation $Z$ does not change according to treament label $T$. In contrast, t
    The mutual information guided disentangled representation proposed in this paper is different to these works in that the target label is regularized to be conditionally independent when the representation and covariate are given.
    
% 2. Local simmilarity preserving 하는 논문들 SITE. NNM, HSIC-NN\\
    Properly exploiting the local structure of each covariate in learning balanced representation is an important issue in the ITE. The most common idea is known as the nearest neighbor matching~(NNM), which finds opponent treatment data for each covariate \cite{johansson16,li2017matching,yao2018}. The predictive variance regularization is also considered to find some kernel in \cite{Alaa17,jean2018semi}. In contrast to these studies, we propose to apply this predictive variance regularization to the learning of stochastic encoders. Recently, \cite{du2019adversarial} also proposed the  balancing method in an adversarial way for causal inference.

    %Properly exploiting the local structure of each covariate with representation balancing is an important issue in the ITE. The most common idea is known as the nearest neighbor matching(NNM), which finds opponent treatment data for each covariate. \cite{johansson16} theoretically showed that nearest neighbor error term exsits in addition to representation balancing term in the error bound of counterfactual risk. In \cite{li2017matching}, NNM is developed through low-dimensional subspace projection for empirical performance. \cite{yao2018} introduces middle-distance distance minimization (MPDM), which balances the globally representation, and position-dependent deep metric (PDDM), which preserves the similarity of the locally representation, based on the idea of metric learning. In the context of causal inference, predictive variance regularization is considered to find a kernel that minimizes the Precision in Estimation of Heterogeneous Effect (PEHE) functionalities \cite{Alaa17}. In contrast to the existing methodologies proposed for kernel learning~\cite{jean2018semi, Alaa17}, we propose to apply this predictive variance regularization to the learning of stochastic encoders. Recently, ~\cite{du2019adversarial} also proposed adversarial balancing method for causal inference. Adversarial balancing method, proposed in \cite{du2019adversarial}, differs from our method in that our CIB tries to minimize mutual information between covariate and representation, while adversarial balancing tries to maximize.
    
\paragraph{Generative models for causal inference}

% 1. VAE와 CEVㅇAE에 대한 언급. Latent variable을 모델링 하며 CEVAE에서는 explicit하게 balanced representation를 고려하지 않음.
    
    The stochastic encoder $ p_{\phi} (z | x) $ of CIB can be understood as a generative model in that it samples representations satisfying the information bottleneck principle for causal inference. The variational information bottleneck \cite{Alemi17} is known as the supervised learning version of $\beta$-VAE~\cite{Higgins17}. Despite their similarities on network architectures and loss functions, they have different interpretations; VIB with stochastic encoders variationally approximates the marginal representation distribution $ p (z) $ and the conditional distribution $ p (y | z) $, but VAE with the prior and sample generators approximates the posterior distribution $ p (z | x)$. This difference is related to regularizations on $z$. Since our model is the natural extension of VIB for causal inference, it inherits this property and has a similar connection to CEVAE \cite{Louizos17}. %In addition, VAE and VIB also differ as to which random variable the encoder should expressive. For VAE representation $Z$ must be expressive for covariate $ X $, which is represented by a reconstruction loss. On the other hand, the VIB should expressive $ Z $ for outcome $ Y $, and only consider compression for covariate $ X $. 
    
     Information bottleneck principle is used for causal inference by recent parallel work \cite{parbhoo2018causal}. A graphical structure for CEIB~\cite{parbhoo2018causal} assumes that the representation $Z$ directly affects to outcome $Y$. On the other hand, CIB naturally extends the standard information bottleneck principle and inherits its benefits. This difference between CIB and CEIB influences the details of architecture selections and objective functions. 
     
     %also proposed information bottleneck principles to perform causal inference. While their method focus on learning a discrete, low-dimensional latent space representation of confounding information, our method models representation as conditional Gaussian distribution. Also, graphical model for CEIB~\cite{parbhoo2018causal} assumes representation $Z$ directly affects to outcome $Y$ while CIB assumes representation $Z$ is conditionally independent of outcome $Y$ given covariate $X$. This difference between CIB and CEIB influences details of architecture selection and objective function. Specifically, CIB uses single encoder for treat/control data, while CEIB uses separate encoders. Also, our mutual information guided disentanglement is not considered in CEIB, since representation in CEIB directly affects to the observed outcome.

\section{Experiment}
%\subsection{Experiments on real-world benchmark datasets}
We evaluate CIB on three real-world datasets used in the existing literature \cite{johansson16, Shalit17, Louizos17}. Due to the space constraints, we defer descriptions on datasets and evaluation metrics to the appendix. 
\begin{table}[t]
\caption{Comparisons of counterfactual errors: 10 repeat/in-sample case}
  \label{metric-in}
  \centering
  {\scriptsize\begin{tabular}{l>{\centering\arraybackslash}p{2.5cm}>{\centering\arraybackslash}p{2.5cm}>{\centering\arraybackslash}p{2.5cm}}
    \toprule
    Dataset & IHDP($\sqrt{\epsilon_{PEHE}}$) & Jobs($\mathcal{R}_{pol}$) & Twins(AUC) \\
    \midrule
    TARNET & $\mathbf{0.729 \pm 0.088}$ & $0.228 \pm 0.004$ & $0.849 \pm 0.002$ \\
    CFR-M & $\mathbf{0.663 \pm 0.068}$ & $0.213 \pm 0.006$ & $0.852 \pm 0.001$ \\
    CFR-W & $\mathbf{0.649 \pm 0.089}$ & $0.225 \pm 0.004$ & $0.850 \pm 0.002$ \\
    CEVAE & $(2.7 \pm 0.1)$ & $\mathbf{0.15 \pm 0.0}$ & not reported\\
    SITE & $\mathbf{0.604 \pm 0.093}$ & $0.224 \pm 0.004$ & $0.862 \pm 0.002 $\\
    \midrule
    CIB & $ \mathbf{0.663 \pm 0.193}$ & $0.256 \pm 0.006$ & $\mathbf{0.870 \pm 0.002}$ \\
    \bottomrule
    %\multicolumn{4}{c}{*Bold font indicates that the mean belongs to 95\% confidence interval of the best performing model.}
  \end{tabular}}
%\end{table}
%\vspace{-1cm}
%\setlength{\abovecaptionskip}{0cm}
%\begin{table}[t]
  \caption{Comparisons of counterfactual errors: 10 repeat/out-sample case}
  \label{metric-out}
  \centering
  {\scriptsize\begin{tabular}{l>{\centering\arraybackslash}p{2.5cm}>{\centering\arraybackslash}p{2.5cm}>{\centering\arraybackslash}p{2.5cm}}
    \toprule
    Dataset & IHDP($\sqrt{\epsilon_{PEHE}}$) & Jobs($\mathcal{R}_{pol}$) & Twins(AUC) \\
    \midrule
    TARNET & $1.342 \pm 0.597$ & $0.234 \pm 0.012$ & $0.840 \pm 0.006$ \\
    CFR-M & $1.202 \pm 0.550$ & $0.231 \pm 0.009$ & $0.840 \pm 0.006$ \\
    CFR-W & $1.152 \pm 0.527$ & $0.225 \pm 0.010$ & $0.842 \pm 0.005$ \\
    CEVAE & $(2.6 \pm 0.1)$ & $ 0.26 \pm 0.0$ & not reported\\
    SITE & $\mathbf{0.656 \pm 0.108}$ & $\mathbf{0.219 \pm 0.009}$ & $\mathbf{0.853 \pm 0.006} $\\
    \midrule
    CIB & $\mathbf{0.613 \pm 0.118}$ & $\mathbf{0.211 \pm 0.017}$ & $\mathbf{0.861 \pm 0.005}$ \\
    \bottomrule
    \multicolumn{4}{l}{*Bold font indicates that the mean belongs to 95\% confidence interval of the best performing model} \\
    \multicolumn{4}{l}{on each dataset.} \\
  \end{tabular}}
  \vspace{-15pt}
\end{table}

% 3. 실험 셋팅 설명
    \vspace{-.1cm}
    \paragraph{Baselines} We compare our method against classical regression methods, nearest neighbor based methods and representation learning based methods. Classical regression methods involve Ordinary Least Squares with treatment as a feature (OLS/LR 1), OLS with separate regressors for each treamtment (OLS-2). Nearest neighbor based methods contain Hilbert-Schmidt Independence Criterion based Nearest Neighbor Matching (HSIC-NNM)~\cite{chang2017informative}, Propensity Score Match with logistic regression (PSM)~\cite{rosenbaum1983central} and $k$-nearest neighbor ($k$-NN)~\cite{crump2008nonparametric}. Representation learning based methods contain Balanced Neural Network (BNN)~\cite{johansson16}, Treatment-Agnostic Representation Network (TARNET), Counterfactual regression with MMD/Wasserstein metric (CFR-M/W)~\cite{Shalit17}, Generative Adversarial Nets for inference of ITE (GANITE)~\cite{yoon2018ganite}, Causal Effect Variational Autoencoder (CEVAE)~\cite{Louizos17} and local Similarity preserved Individual Treatment Effect (SITE)~\cite{yao2018}. We also report the results for SITE with MMD/Wasserstein metric (SITE-M/W). Results in Table \ref{metric-in} and \ref{metric-out} except our model are reported in \cite{yoon2018ganite, Louizos17, yao2018}. We parenthesized the results of CEVAE and GANITE, since their results are based on slightly different setting: 1000 realization for IHDP dataset and 100 realization for Jobs dataset (GANITE).
    
    \vspace{-.1cm}
    \paragraph{Implementation} We use a shared encoder for both treat/control data and separate regressors as in TARNET~\cite{Shalit17}.  Using a Gaussian reparameterization trick as proposed in VIB~\cite{Alemi17}, our stochastic encoder predicts mean and standard deviation of conditional Gaussian given covariates. We use the same number of hidden layers for conditional mutual information estimator and encoder. We use single-layered networks for regressors. The score classifier did not improve empirical performance of CIB, and was omitted. More details on implementations are provided in the appendix. 

% 4. 기존 모델 대비 성능 개선 
    \vspace{-.1cm}
    \paragraph{Results} Tables \ref{metric-in} and \ref{metric-out} represent means and standard errors of 10 realizations/splits on three datasets. CIB achieves the best result for in-sample error on Twins dataset and out-sample error on IHDP and Twins datasets. On Jobs datasets, CIB shows the comparable performance to other state-of-the-art models. Tables here only show comparisons against major baselines and full comparisons are provided in the appendix due to the space constraints. 
    
    As noted in \cite{yao2018}, the results show that the representation learning based methods outperform the classical regression methods and the nearest neighbor based methods. Our CIB method also follows this tendency. Although other representation balancing methods such as BNN~\cite{johansson16}, CFR~\cite{Shalit17} and SITE~\cite{yao2018} consider \emph{group-wise} balancing representation, i.e. their regularization for balancing representation is defined for a group of representations, while MIGDR of CIB considers seeks \emph{instance-wisely} balanced representation. Since CIB adds an auxiliary network for balancing representations, it does not suffer from calculating the Sinkhorn divergence in CFR-W or choosing optimal kernel in CFR-M. Also, CIB does not need to hold hard examples to consider local similarity unlike SITE.
        
    \vspace{-.1cm}
    \paragraph{Uncertainty calibration with information bottleneck penalty}
    
    \begin{wrapfigure}{r}{0.52\textwidth}
        \centering
        \vspace{-20pt}
        \includegraphics[width=0.52\textwidth]{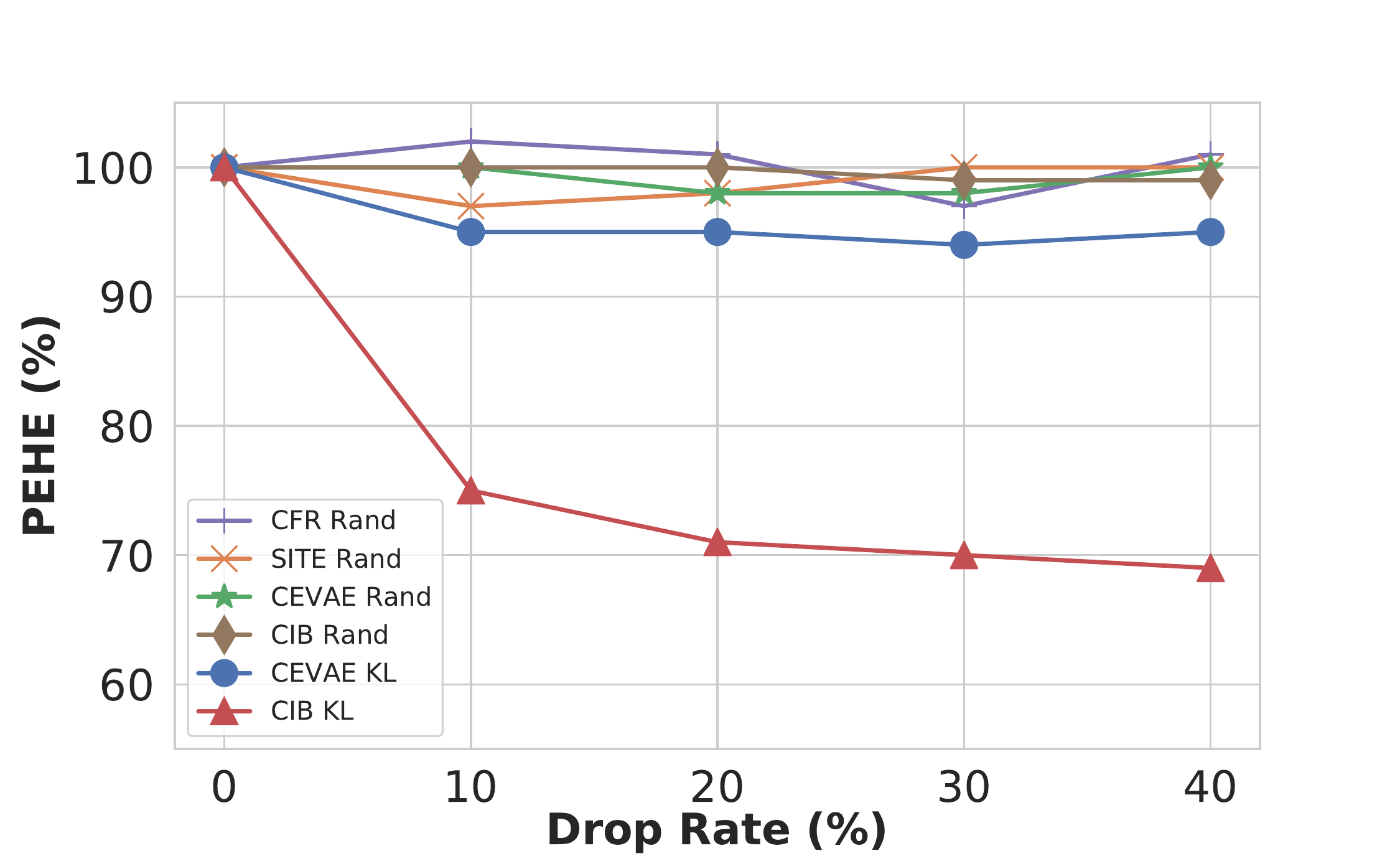}
        \caption{Results on removing top $k\%$ "I don't know" samples }
        \vspace{-12pt}
 	\label{fig:unc-calib}
    \end{wrapfigure}

    Since the representation of CIB is modeled as conditional Gaussian, we can measure how generated representation $z$ given covariate $x$ is rare compared to marginal representation as $\log\frac{p_{\phi}(z|x)}{r_{\psi}(z)}$. Therefore, if we average this with sampled representation as $\mathbb{E}_{p_{\phi}(z|x)}[\log\frac{p_{\phi}(z|x)}{r_{\psi}(z)}]  {=} D_{KL}( p_{\phi}(z|x) \| r_{\psi}(z) )$, we can interpret this term as an indicator for \textit{how much the covariate $x$ is out-of-distribution (OOD) sample} for entire population $p(x)$ \cite{alemi2018}. In order to explicitly benefit this property, we consider the scenario in this experiment where models are allowed to answer ``I don't know'' for uncertain inputs. 
    
    %Hence, we can use this indicator to uncertainty calibrator for evaluation and reduced top $k\%$ uncertain sample in test time. Knowing "I don't know" samples are quite important in practice such as healthcare and political science. Wrong predictions for such samples can be worse than doing nothing for mission-critical application of causal inference. 
    
    In Figure \ref{fig:unc-calib}, we show our results on removing top $k$\% uncertain samples for IHDP datasets. %As baselines, we compare our sample dropping scenario to random drop with baseline methods. 
    Since CEVAE~\cite{Louizos17} can also measure $D_{KL}( p_{\phi}(z|x,t,y) \| r_{\psi}(z) )$ for confounder $Z$, we also remove uncertain samples based on this KL divergence (denoted as "CEVAE KL"). However, since the predictive variances are unavailable for other methods, we just drop $k$\% samples randomly for them. As Figure \ref{fig:graphical_model} shows, only CIB quickly finds and drops samples on which it is not sure.

    %random drop method or KL calibration in CEVAE do not find any prediction degrading sample, while our methods finds quickly drop such samples for only $10\%$ dropping.

    \vspace{.2cm}
    \paragraph{Ablation Study} 
    We proposed two additional regularizations above, MIGDR and CPVR on top of basic CIB framework. Here, we confirm our additional regularization is critical to empirical success of our model. Toward this, we compare our CIB against CIB without MIGDR, CIB without CPVR and CIB with no additional regularization on three datasets. Table \ref{ablation-study} summarizes the results. The CPVR regularization is critical to results for IHDP dataset, while both MIGDR and CPVR are important for other datasets.

% \begin{table}[t]
%   \centering
%   \caption{Ablation Study on In-sample}
%   \label{ablation-study-in}
%   {\scriptsize\begin{tabular}{lccc}
%     \toprule
%     Dataset & IHDP($\sqrt{\epsilon_{PEHE}}$) & Jobs($\mathcal{R}_{pol}$) & Twins(AUC) \\
%     \midrule
%     CIB & ${0.663 \pm 0.193}$ & ${0.256 \pm 0.006}$ & ${0.870 \pm 0.002}$ \\
%     \midrule
%     without MIGDR & $0.664 \pm 0.191$ & $0.246 \pm 0.004$ & $0.864 \pm 0.001$ \\
%     without CPVR & $0.686 \pm 0.186$& $0.245 \pm 0.013$ & $0.865 \pm 0.001$ \\
%     No regularization & $0.686 \pm 0.186$ & $0.245 \pm 0.004$ & $0.865 \pm 0.001$ \\
%     \bottomrule
%   \end{tabular}}
% \end{table}

% \begin{table}[t]
%   \centering
%   \caption{Ablation Study on Out-sample}
%   \label{ablation-study-out}
%   {\scriptsize\begin{tabular}{lccc}
%     \toprule
%     Dataset & IHDP($\sqrt{\epsilon_{PEHE}}$) &Jobs($\mathcal{R}_{pol}$) & Twins(AUC) \\
%     \midrule
%     CIB & ${0.613 \pm 0.118}$ & ${0.211 \pm 0.017}$ & ${0.861 \pm 0.005}$ \\
%     \midrule
%     without MIGDR & $0.614 \pm 0.118$ & $0.230 \pm 0.013$ & $0.858 \pm 0.006$ \\
%     without CPVR & $0.649 \pm 0.122$ & $0.230 \pm 0.013$ & $0.858 \pm 0.006$ \\
%     No regularization & $0.649 \pm 0.122$ & $0.230 \pm 0.013$ & $0.858 \pm 0.006$ \\
%     \bottomrule
%   \end{tabular}}
% \end{table}

\setlength{\textfloatsep}{0.5cm}
\begin{table}[t]
  \centering
  \caption{Results of CIB without MIGDR or CPVR}
  \label{ablation-study}
  {\scriptsize\begin{tabular}{lcccccc}
    \toprule
    \multirow{2}{*}{Dataset} & \multicolumn{2}{c}{IHDP($\sqrt{\epsilon_{PEHE}}$)} & \multicolumn{2}{c}{Jobs($\mathcal{R}_{pol}$)} & \multicolumn{2}{c}{Twins(AUC)} \\
    & In-sample & Out-sample & In-sample & Out-sample & In-sample & Out-sample \\
    \midrule
    CIB & ${0.663 \pm 0.193}$ & ${0.613 \pm 0.118}$ & ${0.256 \pm 0.006}$ & ${0.211 \pm 0.017}$ & ${0.870 \pm 0.002}$ & ${0.861 \pm 0.005}$ \\
    \midrule
    w/o MIGDR & $0.664 \pm 0.191$ & $0.614 \pm 0.118$ & $0.246 \pm 0.004$ & $0.230 \pm 0.013$ & $0.864 \pm 0.001$ & $0.858 \pm 0.006$ \\
    w/o CPVR & $0.686 \pm 0.186$ & $0.649 \pm 0.122$ & $0.245 \pm 0.013$ & $0.230 \pm 0.013$ & $0.865 \pm 0.001$ & $0.858 \pm 0.006$ \\
    No regularizer & $0.686 \pm 0.186$ & $0.649 \pm 0.122$ & $0.245 \pm 0.004$ & $0.230 \pm 0.013$ & $0.865 \pm 0.001$ & $0.858 \pm 0.006$ \\
    \bottomrule
  \end{tabular}}
\end{table}

\section{Conclusion}
    \vspace{-.15cm}
    We introduced a novel framework CIB for estimating ITE by extending information bottleneck. On the top of information bottleneck framework, we proposed two additional regularizations to learn more reliable representation. We confirmed CIB method performs comparable to state-of-the-art models and reliable in detecting \textit{how much given covariate $x$ is out-of-distribution (OOD) sample} for entire population $p(x)$. This property of CIB is critical to mission-critical applications of causal inference.

%\section{Acknowledgments}

%1. Information based technique으로 confounder를 모델링하고 이를 기반으로 예측하는 것의 중요성. 또한, Perturbation consistency loss를 사용하는 것의 중요성.

\bibliographystyle{apa}
\bibliography{cib}

\newpage
\appendix
\section{Details on MGIDR}
    As mentioned in main text, conditional mutual information between $Z$ and $T$ can be represented as
    \begin{align*}
    % \centering
        \mathbb{E}_{p(x)}\big[D_{KL}(p(z,t|x)\|p_{\phi}(z|x)p(t|x))\big]
        =\mathbb{E}_{p(x)}\Big[\sup_{f} \mathbb{E}_{p(z,t|x)}[f] - \log \mathbb{E}_{p_{\phi}(z|x)p(t|x)}[e^{f}]\Big].
    \end{align*}
    where $f : \mathcal{X} \times \mathcal{Z} \times \mathcal{T} \rightarrow \mathbb{R}$ is a scalar-valued function whose two expectation are finite. Due to the universal approximation properties of the neural network, this statistic is strongly consistent when we approximate $ f_{\rho} $ with neural networks without optimizing on the infinite dimensions\cite{Belghazi18}. Learning to maximize the mean difference between the two distributions is similar to learning discriminators in generative adversarial learning. Because it is less important to calculate the exact mutual information in this paper, we optimize the statistic network $f$ using a zero-centered gradient penalty\cite{Thanh19}, which is more stable and known to be capable of greater generalization performance outcomes. Moreover, finding the optimal $f_{\rho}$ for each covariate $x$ is inefficient because doing so requires the estimation of several functions. Therefore, the statistic network is learned with amortized inference by considering $ x $ conditionally.

\section{Details on Experiments}   
\paragraph{Datasets} 
    To evaluate the proposed CIB, we use three real-world datasets used to evaluate the existing methodologies\cite{johansson16, Shalit17, Louizos17}. The covariates of Infant Health and Development Program (IHDP) dataset are from real-world randomized controlled trial experiment. By removing some of the treated observations of RCT dataset, the IHDP dataset intentionally involves selection bias and is used as benchmark data for ITE researches. More details are given in \cite{hill2011bayesian}. The data consists of 25 dimensions 747 covariates . We split train/validate/test 10 realization of simulated data with 63/27/10 ratio as proposed in \cite{yao2018}. Jobs dataset \cite{LaLonde1986} is a mixed observations based on the National Support Work program with observational study \cite{Smith2005}. The covariate of Jobs dataset consists with 17 dimension 3212 instance. Since this dataset consists of one realization without repeated simulation, we experimented with 10 times train/validation/test split with 56/24/20 ratio as suggested in \cite{Shalit17}. The Twins dataset was used as the benchmark data of counterfactual inference in for the first time in \cite{Louizos17}. This data is based on the twins born in the USA in 1989-1991. As in \cite{yao2018}, we focus on the same sex twin-pair less than 2000g. The covariate is a 40-dimensional data consisting of parents, their condition at the time of their pregnancy, and birth information. We used the identical data preprocessing as suggested in \cite{yao2018}.
    \paragraph{Performance Metrics}
    In the case of IHDP datasets where all control/treat outcomes are known for each covariate, we use the expected Precision in Estimation of Heterogeneous Effect (PEHE). Since there exists only factual outcome for a given covariate, it is intractable to calculate the PEHE score in Jobs data. In order to evaluate the performance of the learned model, \cite{Shalit17} defined a policy $\pi_f$ based on the learned model $f$ as follows.
    \begin{equation}
        \pi_f(x) = \begin{cases}1, \text{ if }f(x,0) \le f(x,1)\\
        0, \text{ if }f(x,0) > f(x,1)
        \end{cases}
    \end{equation}
    \begin{equation}
     \mathcal{R}_{pol}(\pi_f) = 1- \Big( \mathbb{E}[Y_1|\pi_f(x)=1]\cdot p(\pi_f=1) + \mathbb{E}[Y_0|\pi_f(x)=0]\cdot p(\pi_f=0) \Big)
    \end{equation}
    We estimate this for the randomized subset of Jobs dataset as
    \begin{equation}
        \hat{\mathcal{R}}_{pol} = 1 - \Big( \frac{1}{|X_1 \cap T_1 \cap E|}\sum_{x_i \in X_1 \cap T_1 \cap E} y_1^{(i)} \frac{|X_1 \cap E|}{|E|} + \frac{1}{|X_0 \cap T_0 \cap E|}\sum_{x_i \in X_0 \cap T_0 \cap E} y_0^{(i)} \frac{|X_0 \cap E|}{|E|}  \Big)
    \end{equation}
    where $E=\{x_i : x_i \text{ is from the randomized experiment.}\}$, $X_1 = \{ x_i : \hat{y}_1^{(i)} - \hat{y}_0^{(i)} > 0 \}$, $A_0 = \{ x_i : \hat{y}_1^{(i)} - \hat{y}_0^{(i)} \le 0 \}$,$T_1 = \{ x_i : t_i =1 \}$, $T_0 = \{ x_i : t_i =0 \}$.
    Finally, for the Twins dataset, we report area under ROC (AUC) for counterfactual prediction as proposed in \cite{Louizos17}.
    
    We measure the in-sample error with both train and valid data and out-sample error with only test data. This convention follows \cite{johansson16, Shalit17, yao2018}. Since in-sample evaluation metric is non-trivial, as we never observe the ITE for any unit, we report in-sample results.
    
    We measure the in-sample error with both train and valid data and out-sample error with only test data. This convention follows \cite{johansson16, Shalit17, yao2018}. Since in-sample evaluation metric is non-trivial, as we never observe the ITE for any unit, we report in-sample results.
    
    \paragraph{Implementation} For cross-validation, We search to optimize the number of hidden layers in encoder ($\{1,2,3\}$), the dimension of layers ($\{64, 128, 256 \})$ and the regularization coefficients ($\beta, \lambda_{M}, \lambda_{V} \in \{ 0.01, 0.1, 1.0, 10.0, 100.0\}$). We use Adam\cite{Kingma2015} with $0.0001$ for learning rate and $\beta_1 = 0.9, \beta_2 = 0.999$. We train 2000 iterations with early stopping as \cite{Shalit17, yao2018}. Experiments in this paper are conducted on a cluster with Intel Xeon E5 2.2GHz CPU, 4x Nvidia Titan Xp GPU and 256GB RAM.
    
\section{Full Experiment Result}
Bold font indicates that the mean belongs to 95\% confidence interval of the best performing model on each dataset.
\begin{table}[ht]
  \caption{Mean/Std Err for counterfactual error 10 repeat/in-sample}
  \label{metric-in_full}
  \centering
  \begin{tabular}{lccc}
    \toprule
    Dataset & IHDP($\sqrt{\epsilon_{PEHE}}$) & Jobs($\mathcal{R}_{pol}$) & Twins(AUC) \\
    \midrule
    OSL/LR1 & $10.761 \pm 4.350$ & $0.310 \pm 0.017$ & $0.660 \pm 0.005$ \\
    OSL/LR2 & $10.280 \pm 3.794$ & $0.228 \pm 0.012$ & $0.660 \pm 0.004$ \\
    \midrule
    HSIC-NNM & $2.439 \pm 0.445$ & $0.291 \pm 0.019$ & $0.762 \pm 0.011$ \\
    PSM & $7.188 \pm 2.679$ & $0.292 \pm 0.019$ & $0.500 \pm 0.003$ \\
    $k$-NN & $4.432 \pm 2.345$ & $0.230 \pm 0.016$ & $0.609 \pm 0.010$ \\
    \midrule
    BNN & $3.827 \pm 2.044$ & $0.232 \pm 0.008$ & $0.690 \pm 0.008$ \\
    TARNET & $\mathbf{0.729 \pm 0.088}$ & $0.228 \pm 0.004$ & $0.849 \pm 0.002$ \\
    CFR-M & $\mathbf{0.663 \pm 0.068}$ & $0.213 \pm 0.006$ & $0.852 \pm 0.001$ \\
    CFR-W & $\mathbf{0.649 \pm 0.089}$ & $0.225 \pm 0.004$ & $0.850 \pm 0.002$ \\
    GANITE & $(1.9 \pm 0.4)$ & $({0.13 \pm 0.01})$ & not reported\\
    CEVAE & $(2.7 \pm 0.1)$ & $\mathbf{0.15 \pm 0.0}$ & not reported\\
    SITE-M & $1.162 \pm 0.118$ & $0.194 \pm 0.015$ & $0.710 \pm 0.003$ \\
    SITE-W & $0.993 \pm 0.112$ & $0.190 \pm 0.015$ & $0.849 \pm 0.003$ \\
    SITE & $\mathbf{0.604 \pm 0.093}$ & $0.224 \pm 0.004$ & $0.862 \pm 0.002 $\\
    \midrule
    CIB & $ \mathbf{0.663 \pm 0.193}$ & $0.256 \pm 0.006$ & $\mathbf{0.870 \pm 0.002}$ \\
    \bottomrule
  \end{tabular}
\end{table}

\begin{table}[ht]
  \caption{Mean/Std Err for counterfactual error 10 repeat/out-sample}
  \label{metric-out_full}
  \centering
  \begin{tabular}{lccc}
    \toprule
    Dataset & IHDP($\sqrt{\epsilon_{PEHE}}$) & Jobs($\mathcal{R}_{pol}$) & Twins(AUC) \\
    \midrule
    OSL/LR1 & $7.354 \pm 2.914$ & $0.279 \pm 0.067$ & $0.500 \pm 0.028$ \\
    OSL/LR2 & $5.245 \pm 0.986$ & $0.733 \pm 0.103$ & $0.500 \pm 0.016$ \\
    \midrule
    HSIC-NNM & $2.401 \pm 0.367$ & $0.311 \pm 0.069$ & $0.501 \pm 0.017$ \\
    PSM & $7.290 \pm 3.389$ & $0.307 \pm 0.053$ & $0.506 \pm 0.011$ \\
    $k$-NN & $4.303 \pm 2.077$ & $0.262 \pm 0.038$ & $0.492 \pm 0.012$ \\
    \midrule
    BNN & $4.874 \pm 2.850$ & $\mathbf{0.240 \pm 0.012}$ & $0.676 \pm 0.008$ \\
    TARNET & $1.342 \pm 0.597$ & $\mathbf{0.234 \pm 0.012}$ & $0.840 \pm 0.006$ \\
    CFR-M & $1.202 \pm 0.550$ & $\mathbf{0.231 \pm 0.009}$ & $0.840 \pm 0.006$ \\
    CFR-W & $1.152 \pm 0.527$ & $\mathbf{0.225 \pm 0.010}$ & $0.842 \pm 0.005$ \\
    GANITE & $(2.4 \pm 0.4)$ & $({0.14 \pm 0.01})$ & not reported\\
    CEVAE & $(2.6 \pm 0.1)$ & $ 0.26 \pm 0.0$ & not reported\\
    SITE-M & $1.242 \pm 0.153$ & $\mathbf{0.218 \pm 0.010}$ & $0.705 \pm 0.006$ \\
    SITE-W & $1.459 \pm 0.481$ & $\mathbf{0.232 \pm 0.011}$ & $0.762 \pm 0.007$ \\
    SITE & $\mathbf{0.656 \pm 0.108}$ & $\mathbf{0.219 \pm 0.009}$ & $\mathbf{0.853 \pm 0.006}$\\
    \midrule
    CIB & $\mathbf{0.613 \pm 0.118}$ & $\mathbf{0.211 \pm 0.017}$ & $\mathbf{0.861 \pm 0.005}$ \\
    \bottomrule
    % \multicolumn{4}{c}{*Bold font indicates that the mean belongs to 95\% confidence interval of the best performing model.}\\
  \end{tabular}
\end{table}

\end{document}